\documentclass[a4paper]{article}

\usepackage{INTERSPEECH2018}
\usepackage{amsmath,graphicx}
\usepackage{mathtools}
\usepackage{multirow}
\usepackage{hhline}
\usepackage{arydshln}

\DeclarePairedDelimiter\norm{\lVert}{\rVert}%

\title{Memory Time Span in LSTMs for Multi-Speaker Source Separation \thanks{This work was funded by the SB PhD grant of the Research Foundation Flanders (FWO) with project number 1S66217N and the KULeuven research grant GOA/14/005 (CAMETRON).}}
\name{Jeroen Zegers, Hugo Van hamme}
\address{
  KU Leuven, Dept. ESAT, Belgium}
\email{jeroen.zegers@esat.kuleuven.be, hugo.vanhamme@esat.kuleuven.be}

\begin{document}

\maketitle
\begin{abstract}
With deep learning approaches becoming state-of-the-art in many speech (as well as non-speech) related machine learning tasks, efforts are being taken to delve into the neural networks which are often considered as a black box. In this paper it is analyzed how recurrent neural network (RNNs) cope with temporal dependencies by determining the relevant memory time span in a long short-term memory (LSTM) cell. This is done by leaking the state variable with a controlled lifetime and evaluating the task performance. This technique can be used for any task to estimate the time span the LSTM exploits in that specific scenario. The focus in this paper is on the task of separating speakers from overlapping speech. We discern two effects: A long term effect, probably due to speaker characterization and a short term effect, probably exploiting phone-size formant tracks.

\end{abstract}

\section{Introduction}
Deep learning based methods are being used in many fields of speech processing. Standard feed forward neural networks cannot take into account temporal information (unless a context window is used), which seems undesirable for a speech signal. Therefor RNNs are often used which are theoretically able to memorize any input from the past. However, in practice it is found that it is difficult for the standard RNN to be trained such that it can use information from many time steps back because of phenomena known as vanishing and exploding gradients. The LSTM cell was introduced to counter these problems by using a fixed error flow with gates that control the input and output behavior of this flow \cite{hochreiter1997long}. LSTMs have been found to have good memory capabilities for longer time spans \cite{gers1999learning} and are currently being considered state of the art for many speech related tasks \cite{chen2017long,meng2017deep,kim2018sound,zen2015unidirectional,xiong2017microsoft}. 

Even though it is known that LSTMs cope better with long term dependencies than standard RNNs, little research has been done to determine the memory time span for a particular task. In this paper part of the cell's memory will be leaked away on purpose to see how performance depends on lost information from the past. A related approach has been taken for the task of action detection in video \cite{singh2016multi}. A network with an LSTM component was trained and at test time the memory was cleared $k$ time steps prior to the current time step. By evaluating the performance of the network for different $k$ the authors roughly found the relevant time span before performance started degrading. The approach proposed in this paper differs in two ways from there approach. Firstly, the memory forgetting technique is applied during both training and testing. This prevents unexpected behavior due to mismatch (e.g. the network could behave strangely if values are changed at test time while it never happened at training time). It was indeed found that leaking all the memory only at test time gave much worse results than applying full leakage both during training and testing. Furthermore, the network might rely on information in the long term memory that it could also extract from a shorter time span. By applying the memory leakage during training as well, the network could learn to extract this information in the shorter time span. Secondly, in this paper, memory leakage is applied by limiting the output of the forget gate to a number smaller than 1 such that each time step, part of the information in the memory is lost and over time the past is forgotten. In \cite{singh2016multi} at time $t$ the network uses a memory cell state $c$ that was set to 0 at time $t-k$. Therefore $k$ different cell states have to be stored that are each reset at a different time. In experiments for this paper, $k$ could exceed a hundred, which would lead to computational problems, specifically during the back-propagation step in training. The leaking approach used in this paper preserves only a single cell state and is thus computationally more elegant, but time spans in this exponential decay approach are less precise. 

The memory time span of the LSTM will be evaluated on the task of single channel multi-speaker source separation (MSSS). The original sound signals of multiple speakers have to be retrieved from a mixture with overlapping speech, without any prior knowledge of the speakers. This is thus an intraclass separation problem which forces the network to memorize some sort of speaker representation in order to consistently assign part of the mixture to the same speaker. It was even shown that explicitly presenting speaker representations at the network inputs could slightly improve MSSS accuracy \cite{zegers2017improving}. It is expected that heavy memory leakage will hinder the network to retain this representation, making this tasks well suited to study long term dependencies. Common deep learning approaches for MSSS are Deep Clustering (DC) \cite{hershey2016deep,isik2016single}, utterance-level Permutation Invariant Training (uPIT) \cite{Yu2017permutation} and Deep Attractor Networks (DANet) \cite{chen2017deep}. Experiments in this paper will be done using DC but results for other approaches are expected to be similar. 

The remainder of this paper is organized as follows. In section \ref{sec:MSSS} the DC approach to the MSSS task is explained, as well as the importance of speaker representations for the task. The method to leak memory will be described in section \ref{sec:DLeakyLSTM}. Experiments will be discussed in section \ref{sec:exp} and finally a conclusion is given in section \ref{sec:concl}.

\section{Multi Speaker Source Separation}
\label{sec:MSSS}
\subsection{Deep Clustering}
\label{sec:DC}
A mixture of $S$ speakers is generally given by $\mathbf{y}[n]=\sum_{s=1}^{S}\mathbf{x}_s[n]$ where $\mathbf{x}_s[n]$ is the source signal for the $s^{th}$ speaker as recorded by the microphone. For the task of MSSS, signals $\hat{\mathbf{x}}_s[n]$ have to be estimated to be as close as possible to the original $\mathbf{x}_s[n]$. In the time-frequency domain, the same task can be expressed using the Short Time Fourier Transform (STFT) of the signals. The task is then to estimate $\hat{\mathbf{X}}_s(t,f)$ from $\mathbf{Y}(t,f)=\sum_{s=1}^{S}\mathbf{X}_s(t,f)$. Usually, a mask $\hat{\mathbf{M}}_s(t,f)$ is estimated for the $s^{th}$ speaker such that 
\begin{equation}
\label{eq:spec_est}
\hat{\mathbf{X}}_s(t,f)=\hat{\mathbf{M}}_s(t,f)\circ\mathbf{Y}(t,f)
\end{equation}
for every time frame $t=0, \ldots, T$ and every frequency $f=0, \ldots, F$ and with $\circ$ the Hadamard product \cite{Yu2017permutation}. The masks are constrained by $\hat{\mathbf{M}}_s(t,f)\ge 0$ and $\sum_{s=1}^{S}\hat{\mathbf{M}}_s(t,f)=1$ for every time-frequency bin $(t,f)$.

In DC a $D$-dimensional embedding vector $\mathbf{v}_{tf}$ is found for every $(t,f)$ via a mapping $\mathbf{v}_{tf}=f_{\theta}(\mathbf{Y})$. $f_{\theta}$ is based on a deep neural network and is chosen such that $\mathbf{v}_{tf}$ is normalized to unit length. The embedding vectors for every bin are stored as rows in a ($TF\times D$)-dimensional matrix $\mathbf{V}$. A ($TF\times S$)-dimensional target matrix $\mathbf{U}$ is defined such that $u_{tf,s}=1$ if target speaker $s$ has the most energy in the mixture for $(t,f)$ and $u_{tf,s}=0$ otherwise. A permutation independent loss function (the columns in $\mathbf{U}$ can be interchanged without changing the loss function) is then stated as
\begin{equation}
\label{eq:dc_loss}
\begin{split}
\mathcal{L}_{\theta} &= \norm{\mathbf{V}\mathbf{V}^T-\mathbf{U}\mathbf{U}^T}_F^2 \\
&= \sum_{t_1,f_1,t_2,f_2}(\langle \mathbf{v}_{t_1f_1},\mathbf{v}_{t_2f_2}\rangle-\langle \mathbf{u}_{t_1f_1},\mathbf{u}_{t_2f_2}\rangle)^2
\end{split}
\end{equation}
where $\norm{.}_F^2$ is the squared Frobenius norm. 
Afterwards, all embedding vectors are clustered into $S$ clusters $c$ using K-means. The masks are then constructed as follows
\begin{equation}
\hat{\mathbf{M}}_{s,tf}=
\begin{cases}
      1, & \text{if}\ \mathbf{v}_{tf} \in c_s \\
      0, & \text{otherwise}
    \end{cases}.
\end{equation}
Equation \ref{eq:spec_est} can then be used to estimate the original source signals via the inverse STFT and overlap-add \cite{rabiner1975theory}.

\subsection{Speaker representation}
A speaker representation that is often used for speaker identification tasks is the i-vector \cite{dehak2011front,glembek2011simplification}. If such i-vectors are explicitly presented to the input of the DC network, as was done in \cite{zegers2017improving}, possibly less information would have to be retained in the LSTM memory.
To obtain such an i-vector, first a Universal Background Model - Gaussian Mixture Model (UBM-GMM) is trained on development data. A supervector $M$ is derived for the utterance, using the UBM. $M$ is then represented by an i-vector $w$ and its projection based on the total variability space,
\begin{equation}
M=m+Tw,
\end{equation}
where $m$ is the UBM mean supervector, $w$ is the total variability factor or i-vector and $T$ is a low-rank matrix spanning a subspace with important variability in the mean supervector space and is trained on development data \cite{dehak2011front,glembek2011simplification}.

\section{Deep Leaky LSTM}
\label{sec:DLeakyLSTM}
A leaky LSTM is an LSTM that is designed to leak cell state memory. First, the architecture of the regular LSTM is explained. Secondly, the adaptation to make it a leaky LSTM is described and how the amount of leakage can be interpreted in memory time span. Finally, the memory flow in a deep network with LSTMs and how this flow can be controlled are described.

\subsection{The regular LSTM}

\begin{figure}[t]
  \vspace*{-4mm}
  \centering
  \includegraphics[width=\linewidth]{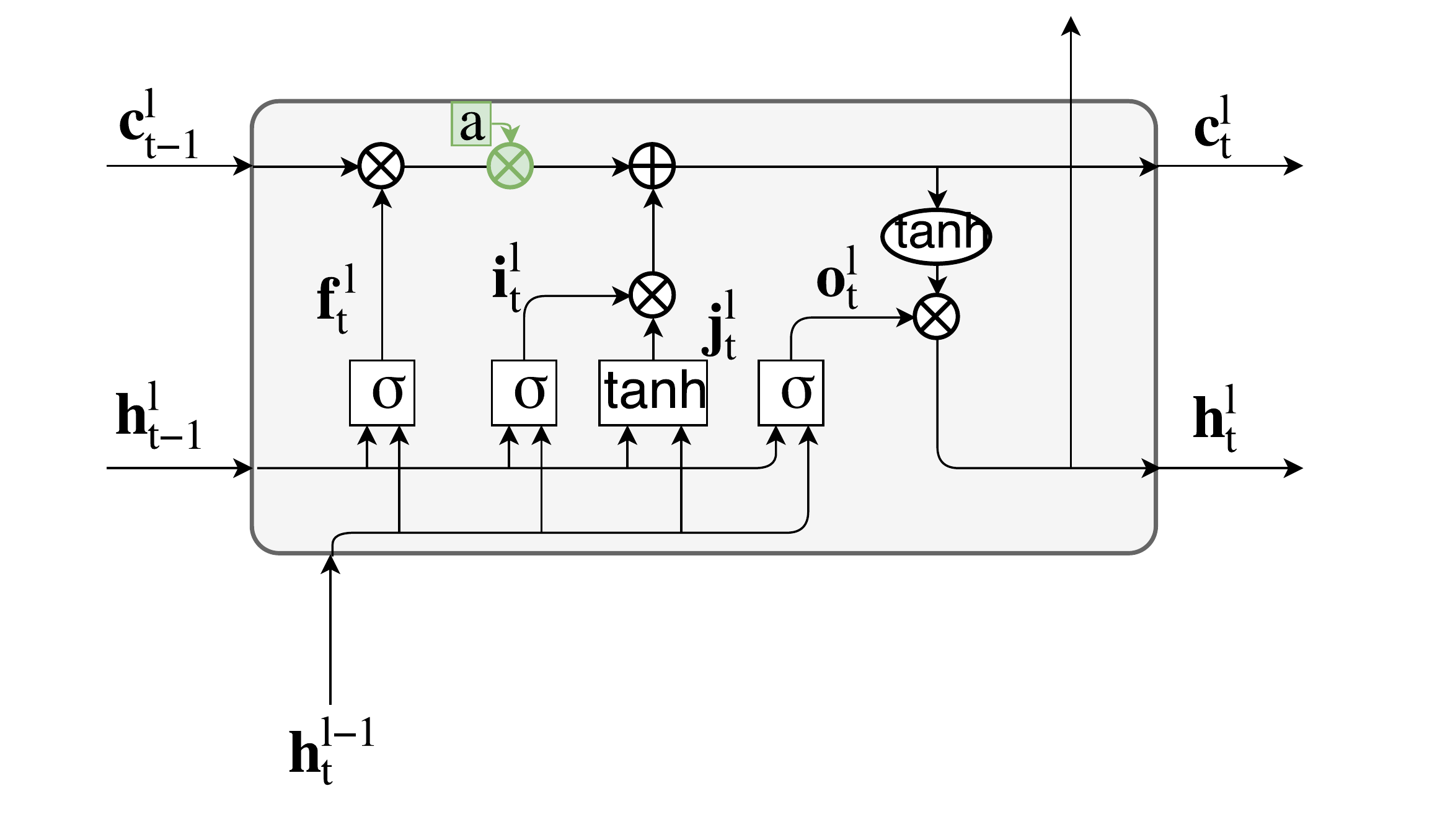}
  \caption{Schematic of a leaky LSTM. Based on \cite{blog}.}
  \label{fig:leakyLSTM}
  \vspace*{-4mm}
\end{figure}

A typical LSTM cell in a layer $l$ of the network is defined as

\begin{equation}
\label{eq:LSTM_f}
\mathbf{f}_t^l=\sigma(\mathbf{W}_f^l \mathbf{h}_t^{l-1} + \mathbf{R}_f^l \mathbf{h}_{t-1}^l + \mathbf{b}_f)
\end{equation}
\begin{equation}
\label{eq:LSTM_i}
\mathbf{i}_t^l=\sigma(\mathbf{W}_i^l \mathbf{h}_t^{l-1} + \mathbf{R}_i^l \mathbf{h}_{t-1}^l + \mathbf{b}_i)
\end{equation}
\begin{equation}
\label{eq:LSTM_o}
\mathbf{o}_t^l=\sigma(\mathbf{W}_o^l \mathbf{h}_t^{l-1} + \mathbf{R}_o^l \mathbf{h}_{t-1}^l + \mathbf{b}_o)
\end{equation}
\begin{equation}
\label{eq:LSTM_j}
\mathbf{j}_t^l=\tanh(\mathbf{W}_j^l \mathbf{h}_t^{l-1} + \mathbf{R}_j^l \mathbf{h}_{t-1}^l + \mathbf{b}_j)
\end{equation}
\begin{equation}
\label{eq:LSTM_c}
\mathbf{c}_t^l=\mathbf{c}_{t-1}^l \odot \mathbf{f}_t^l + \mathbf{j}_t^l \odot \mathbf{i}_t^l
\end{equation}
\begin{equation}
\label{eq:LSTM_h}
\mathbf{h}_t^l= \tanh(\mathbf{c}_{t}^l) \odot \mathbf{o}_t^l
\end{equation}
with $\mathbf{h}_t^{0}=\mathbf{y}_t$, where $\mathbf{y}_t$ is the network's input at time $t$ and $\mathbf{f}_t^l$, $\mathbf{i}_t^l$ and $\mathbf{o}_t^l$ are called the forget gate, the input gate and the output gate, respectively. $\mathbf{c}_t^l$ is called the cell state, $\mathbf{h}_t^{l-1}$ is the cells input and $\mathbf{h}_t^l$ is the cell's output. $\mathbf{W}^l_{*}$ and $\mathbf{R}^l_{*}$ represent trainable weights and $\mathbf{b}_{*}$ trainable biases. These equations are visualized in figure \ref{fig:leakyLSTM} when the green part is ignored.

\subsection{The leaky LSTM}

\begin{figure*}[t]
  \vspace*{-7mm}
  \centering
  \includegraphics[width=\linewidth]{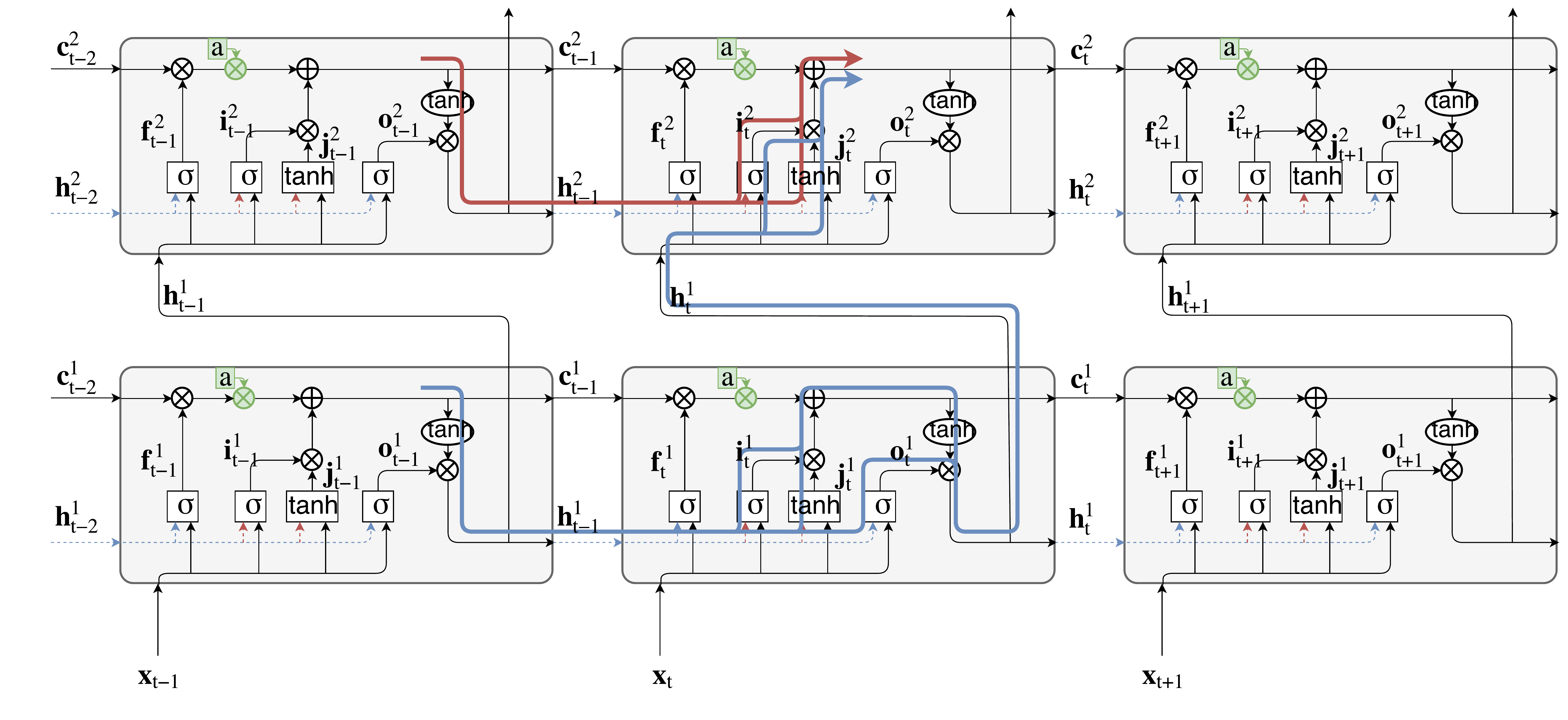}
  \caption{Schematic of flow of temporal information in a deep leaky LSTM and how it can be interrupted by cutting connections. The red line show how recurrent information can bypass the leakage and how this can be countered by removing the red dashed connections. Similar for blue.}
  \label{fig:leakyDLSTM}
  \vspace*{-5mm}
\end{figure*}

A leaky LSTM is a regular LSTM where the forget gate is multiplied by a positive constant $a$ smaller than 1 such that not all cell state information from the previous time steps can be retained and thus part of the memory is leaked. (\ref{eq:LSTM_c}) changes to
\begin{equation}
\label{eq:leaky_LSTM_c}
\mathbf{c}_t^l=\mathbf{c}_{t-1}^l \odot \mathbf{f}_t^l * a + \mathbf{j}_t^l \odot \mathbf{i}_t^l
\end{equation}
as is indicated in figure \ref{fig:leakyLSTM}.

In (\ref{eq:leaky_LSTM_c}) only the first term seems to contain temporal information since only $\mathbf{c}_{t-1}^l$ has the ${t-1}$ subscript. The component of the cell state containing temporal information from time $t-1$ can thus be described as

\begin{equation}
\label{eq:leaky_LSTM_c_fromt-1}
\bar{\mathbf{c}}_{t,t-1}^l= \mathbf{c}_{t-1}^l \odot \mathbf{f}_t^l * a
\end{equation}
The maximal information transfer will be achieved when $\mathbf{f}_t^l=\mathbf{1}$. In this case the contribution from time $t-T$ in the current cell state is given by

\begin{align}
\label{eq:leaky_LSTM_c_fromt-T_expdecay}
\begin{split}
    \bar{\mathbf{c}}_{t,t-T}^l&= \mathbf{c}_{t-T}^l * a^T \\
    &= \mathbf{c}_{t-T}^l * e^{T/\tau}
\end{split}
\end{align}
where the last equality is written in exponential decay form with $\tau=-1/\log(a)$. $\tau$ is called the lifetime and when $T=\tau$, 69\% of the initial value has been lost, which can be considered as the time span for the memory.

\subsection{Temporal information flow}
\label{sec:flow}
It was claimed that temporal information in (\ref{eq:leaky_LSTM_c}) was only present in the $\bar{\mathbf{c}}_{t,t-1}^l$ term. However, the second term in this equation has two factors that indirectly depend on previous time steps via (\ref{eq:LSTM_i}) and (\ref{eq:LSTM_j}). For instance, if $\mathbf{c}_{t-1}^l=\mathbf{1}$, $\mathbf{f}_{t}^l=\mathbf{1}$ and the network would like $\mathbf{c}_{t}^l=\mathbf{c}_{t-1}^l + \Delta \mathbf{c}_{t}^l$, it could obtains this via (\ref{eq:leaky_LSTM_c}) by making $\mathbf{j}_t^l \odot \mathbf{i}_t^l = (1-a) + \Delta \mathbf{c}_t^l$ (with the restriction that $\mathbf{-1} \leqslant \mathbf{j}_t^l \odot \mathbf{i}_t^l \leqslant \mathbf{1}$). This information flow is indicated with the red arrow in figure \ref{fig:leakyDLSTM}. 
This is an unwanted flow since it bypasses the memory leakage. This flow can be stopped by cutting the red dotted connections and thus removing the temporal information in $\mathbf{j}_t^l$ and $\mathbf{i}_t^l$. Equations \ref{eq:LSTM_i} and \ref{eq:LSTM_j} then become

\begin{equation}
\label{eq:leaky_LSTM_i}
\mathbf{i}_t^l=\sigma(\mathbf{W}_i^l \mathbf{h}_t^{l-1} + \mathbf{b}_i)
\end{equation}
\begin{equation}
\label{eq:leaky_LSTM_j}
\mathbf{j}_t^l=\tanh(\mathbf{W}_j^l \mathbf{h}_t^{l-1} + \mathbf{b}_j)
\end{equation}
respectively.

However, in these equations $\mathbf{h}_t^{l-1}$ can also contain temporal information for $l \geqslant 2$ via (\ref{eq:LSTM_h}) and (\ref{eq:LSTM_i},\ref{eq:LSTM_j},\ref{eq:LSTM_o},\ref{eq:LSTM_c}) as is indicated by the blue arrow on figure \ref{fig:leakyDLSTM}. This second order flow can be stopped by also cutting the blue dotted connections. Equations \ref{eq:leaky_LSTM_i} and \ref{eq:leaky_LSTM_j} are retained and  equations \ref{eq:LSTM_f} and \ref{eq:LSTM_o} become

\begin{equation}
\label{eq:leaky_LSTM_f}
\mathbf{f}_t^l=\sigma(\mathbf{W}_f^l \mathbf{h}_t^{l-1} + \mathbf{b}_f)
\end{equation}
\begin{equation}
\label{eq:leaky_LSTM_o}
\mathbf{o}_t^l=\sigma(\mathbf{W}_o^l \mathbf{h}_t^{l-1} + \mathbf{b}_o)
\end{equation}
respectively.

Memory leakage has only been considered on $\mathbf{c}_{t}^l$ while a standard RNN has no cell state and is also able to memorize (even though less effective) via its cell's output $\mathbf{h}_t^{l}$. Cutting the blue and red connection also removes this temporal flow as $\mathbf{h}_{t-1}^{l}$ is never used.

Note that the cutting of recurrent connections is only done to gain more control over the memory leaking process to better define the memory time span. It is not intended to gain performance for the presented task. The research question could be generalized to exploring temporal information used in RNNs with LSTM-like cells.


\section{Experiments}
\label{sec:exp}


\begin{figure*}[t]
  \centering
  \includegraphics[width=\linewidth]{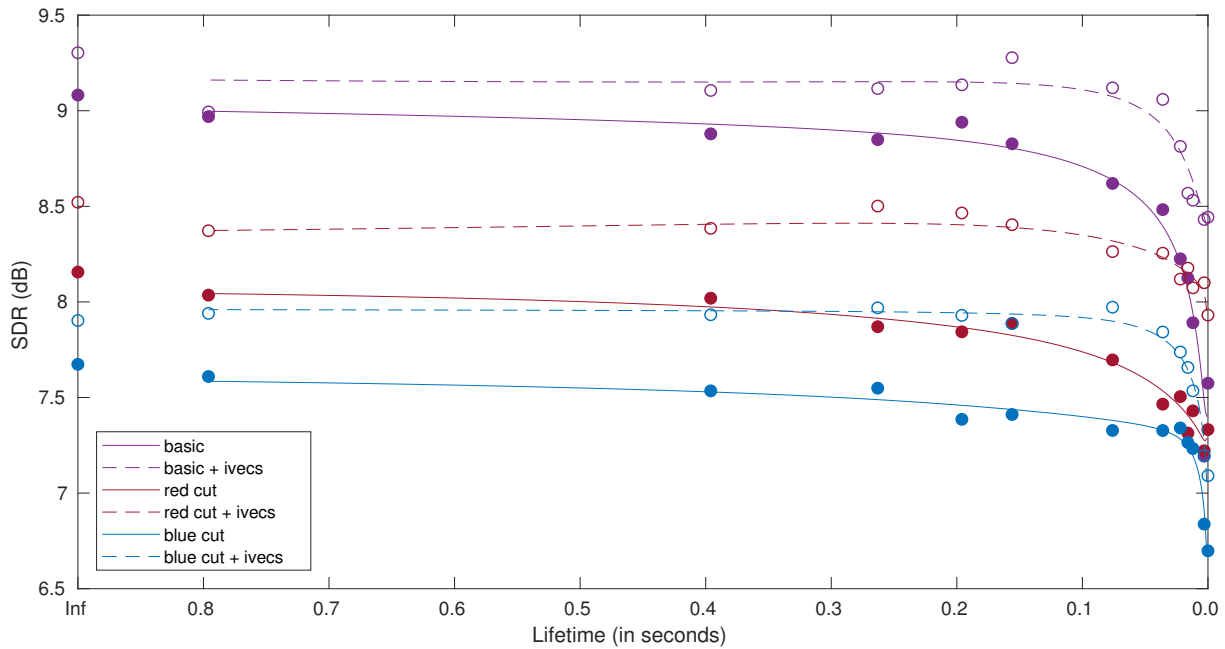}
  \vspace*{-7mm}
  \caption{MSSS results with LSTM leakage. The lifetime is given by $\tau=-1/log(a)*w_{hop}$. Experiments for $\tau=8.0 \, \textrm{s}$ and $\tau=1.6 \, \textrm{s}$ are not plotted but are used to fit the curves.}
  \label{fig:results}
  \vspace*{-3mm}
\end{figure*}

\subsection{Experimental setup}
All experiments were done on two speaker mixtures using the corpus introduced in \cite{hershey2016deep}, which contains artificial mixtures created by mixing together single speaker utterances from the Wall Street Journal 0 (WSJ0) corpus. For every utterance a gain factor was randomly chosen between 0 and 5 dB and utterances were sampled at 8kHz. The length of the mixture was chosen equal to the shortest utterance in the mixture as to maximize the overlap. The training and validation sets contained 20,000 and 5,000 mixtures, respectively and were taken from the \texttt{si\_tr\_s} set of WSJ0. The test set contained 3,000 mixtures using 16 held-out speakers of the \texttt{si\_dt\_05} and \texttt{si\_et\_05} set. The decimal log-magnitude (floored at -300) of the STFT with a 32 ms window length and a hop size ($w_{hop}$) of 8 ms were used as features and were normalized with mean and variance, calculated over the whole training set.

The used network has two fully connected leaky bidirectional LSTM (leaky BLSTM) layers with 600 hidden units each and was trained with the Adam learning algorithm \cite{kingma2014adam}. Curriculum learning was used \cite{bengio2009curriculum}, i.e. the network was presented an easier task before tackling the main task. Here, the network was first trained on 100-frame non-overlapping segments of the mixtures. This network was then used as initialization when training over the full mixture. 
Zero mean Gaussian noise with standard deviation $0.2$ was applied to the training features to avoid local optima. Dropout was not used since it did not improve the results in the experiments. The validation loss was calculated 3 times per epoch and early stopping was applied when the validation loss increased for 9 consecutive times. For DC the embedding dimension was chosen at $D=20$ and since the frequency dimension was $F=129$, the total number of output nodes was $DF=20*129=2580$. Performance for MSSS was measured on the average signal-to-distortion ratio (SDR) improvements on the test set, using the \texttt{bss\_eval} toolbox \cite{vincent2006performance}. All networks were trained using TensorFlow \cite{abadi2016tensorflow} and the code for all the experiments can be found here: \\\texttt{https://github.com/JeroenZegers/Nabu-MSSS}.

To obtain the i-vectors, the UBM and $T$ were trained on development data from the \texttt{si\_tr\_s} set of Wall Street Journal 1 (WSJ1). 13-dimensional Mel-Frequency Cepstral Coefficients (MFCCs) are used as features and a voice activity detector was used to leave out the silence frames. The UBM has 256 Gaussian mixtures and $w$ is 10-dimensional, as was done in \cite{zegers2017improving}. The i-vectors used in the experiments were obtained from the original single speaker utterances of WSJ0 but could also be obtained from speech signal reconstructions after source separation, as was done in \cite{zegers2017improving}. The former was chosen since it provides a cleaner speaker representation. The MATLAB MSR Identity Toolbox v1.0 \cite{sadjadi2013msr} was used to determine the UBM and $T$ and to obtain the i-vectors. 

\subsection{Memory leakage}
Networks were trained and tested for different values of $a$ and for the different architectures as depicted by the dotted cuts in figure \ref{fig:leakyDLSTM}. Some networks were given the oracle i-vectors of both single speech signals used in the mixture. Results are shown in figure \ref{fig:results}.

For all curves a rapid decrease in performance is found for $\tau<100 \, \textrm{ms}$. Here the leaky LSTMs might have difficulty with tracking the formants of the 2 speakers which span roughly $100 \, \textrm{ms}$ \cite{gay1968effect,umeda1975vowel}. For $\tau>100 \, \textrm{ms}$ the networks without i-vectors steadily keep increasing in performance, while the networks with i-vectors have a higher performance which is roughly constant. This seems to indicate that with a bigger time span, the networks without i-vectors manage better to find their own internal speaker representations. The networks with i-vectors don't need to find this speaker representation and therefor there performance stays constant for bigger time span. For no leakage ($\tau=\infty$) an i-vector is still a better speaker representation than the internal representation in the LSTM. When leakage is absolute ($\tau=0$), performance is better when i-vectors are used, indicating that speaker information aids separation if no context is given and assignment is more consistent over frames.

Since the curves for the LSTMs with i-vectors are flat between $\tau=100 \, \textrm{ms}$ and $\tau=300 \, \textrm{ms}$ the network seems to give little importance to phonotactic and lexical information for the task of MSSS. 

The shapes of the curves for all three types of LSTMs are quite similar, indicating that the bypass mechanisms of temporal information flow as described in section \ref{sec:flow} are limited. The performance drop when cutting connections is expected as the model capacity is reduced by removing parameters (see table \ref{tab:par}) and LSTM gates are more difficult to control. 

Experiments were also done where no leakage was applied during training but only during testing. Results differed from those shown in figure \ref{fig:results}. For example, no temporal information can be used if the LSTM with blue cuts and full leakage ($\tau=0$) is used at test time. However, if the network was trained without leakage the SDR was 4.6 dB which is 2.0 dB below the performance obtained when it is also trained with full leakage, although both networks have no memory. This indicates it is indeed necessary to apply leakage also at training time to avoid mismatch.

\begin{table}[t]
  \caption{The number of trainable parameters in the network in millions.}
  \vspace*{-2mm}
  \centering
  \begin{tabular}{c | c c c}
  	& Basic & Red cut & Blue cut \\
    \hline
  $a>0$ &  18.12 & 15.24 &  12.36 \\
  $a=0$	&  15.08 & 12.21 &  10.77
  \end{tabular}
  \label{tab:par}
  \vspace*{-5mm}
  
\end{table}

\section{Conclusions}
\label{sec:concl}
Leakage was applied to the LSTM memory cell to determine the relevant memory time span for MSSS. A short time effect for formant tracking and a long time effect for speaker characterization were found. The same technique could be used to find the relevant time span for other tasks that use LSTMs.


\bibliographystyle{IEEEtran}
\bibliography{mybib}

\end{document}